\documentclass[11pt]{article}

\usepackage[final]{acl}

\usepackage{times}
\usepackage{latexsym}

\usepackage{tipa} 
\usepackage{kotex}
\usepackage{amsmath}
\usepackage{hyperref}
\usepackage{booktabs}
\usepackage{multirow}

\usepackage{xcolor}
\usepackage{colortbl}
\usepackage{algorithm}
\usepackage{algpseudocode}

\usepackage{ifthen}      
\usepackage{pgf} 

\newcommand{\diff}[1]{%
  \pgfmathparse{#1<0 ? 1 : 0}
  \ifnum\pgfmathresult=1
    \scriptsize{\textcolor{red}{(#1)}}%
  \else
    \scriptsize{\textcolor{blue}{(#1)}}%
  \fi
}

\usepackage[T1]{fontenc}

\usepackage[utf8]{inputenc}

\usepackage{microtype}

\usepackage{inconsolata}

\usepackage{graphicx}

\title{HiKE: Hierarchical Evaluation Framework for\\
Korean-English Code-Switching Speech Recognition}

\author{
 \textbf{Gio Paik\textsuperscript{1,5,*}},
 \textbf{Yongbeom Kim\textsuperscript{2,5}},
 \textbf{Soungmin Lee\textsuperscript{3,5}},
 \textbf{Sangmin Ahn\textsuperscript{1,2,5,$\dagger$}},
 \textbf{Chanwoo Kim\textsuperscript{1,4,5,$\dagger$}}
\\\\
 \textsuperscript{1}Theta One AI,
 \textsuperscript{2}Seoul National University,
 \textsuperscript{3}Georgia Institute of Technology,\\
 \textsuperscript{4}Williams College,
 \textsuperscript{5}ROKAF Reserve Forces
\\
 \small{
   \textbf{*Corresponding Author:} \href{mailto:giopaik0@gmail.com}{giopaik0@gmail.com} \quad
   $\dagger$: Equal Contribution
 }
}

\begin{document}

\maketitle
\begin{abstract}
Despite advances in multilingual automatic speech recognition (ASR), code-switching (CS), the mixing of languages within an utterance common in daily speech, remains a severely underexplored challenge. In this paper, we introduce HiKE: the Hierarchical Korean-English code-switching benchmark, the first globally accessible non-synthetic evaluation framework for Korean-English CS, aiming to provide a means for the precise evaluation of multilingual ASR models and to foster research in the field.
The proposed framework not only consists of high-quality, natural CS data across various topics, but also provides meticulous loanword labels and a hierarchical CS-level labeling scheme (word, phrase, and sentence) that together enable a systematic evaluation of a model's ability to handle each distinct level of code-switching. Through evaluations of diverse multilingual ASR models and fine-tuning experiments, this paper demonstrates that although most multilingual ASR models initially exhibit inadequate CS-ASR performance, this capability can be enabled through fine-tuning with synthetic CS data.
HiKE is available at \url{https://github.com/ThetaOne-AI/HiKE}.

\end{abstract}

\section{Introduction}
\label{sec:intro}

Recent advances in Automatic Speech Recognition (ASR) \cite{whisper, canary, granite_speech} have pushed error rates below $5\%$ on standard monolingual ASR benchmarks \cite{Librispeech, FLEURS, TEDLIUM}, enabling novel applications such as vibe coding, AI-assisted language education and automated podcast summarization. These advancements are fundamentally redefining human-computer interaction.
However, it remains a significantly underexplored question whether the performance of these ASR models in monolingual settings can be extended to Code-Switching (CS) scenarios, where multiple languages are mixed within a single utterance. Consequently, the field of CS-ASR remains underdeveloped~\cite{survey}, especially for language pairs involving low-resource and typologically distant languages such as Korean and English. 
This research gap significantly impairs the user experience for the large global population of multilingual individuals who use CS as a natural, everyday part of communication, particularly in regions where English is not the primary language.

\begin{figure}
    \centering
    \includegraphics[width=\linewidth]{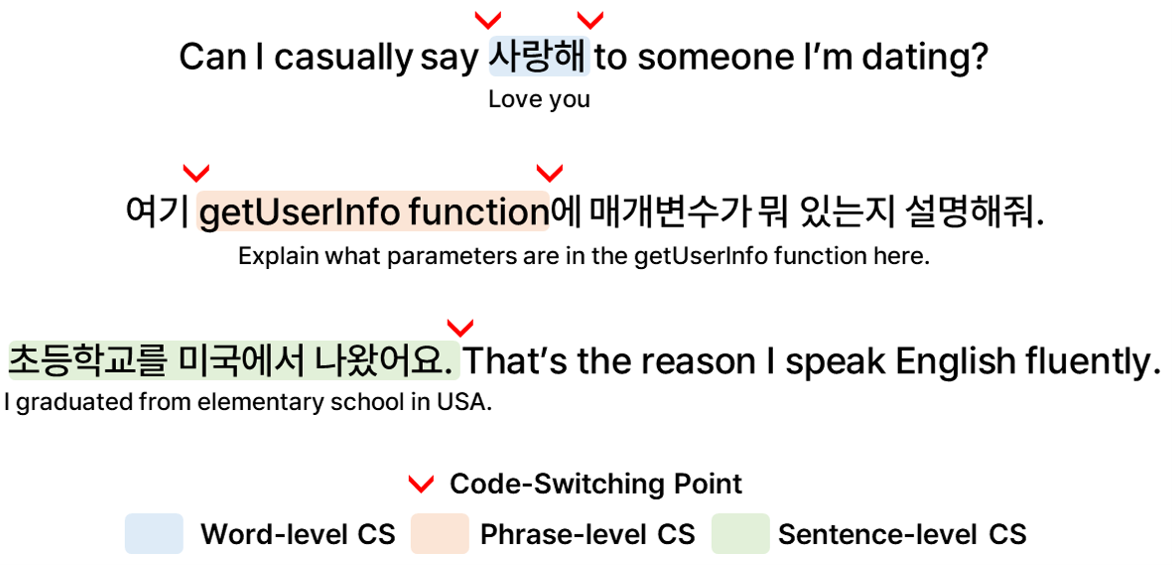}
    \caption{\textbf{Code-Switching Examples by CS-Level}} 
    \label{fig:motivation}
\end{figure}

To address this research gap, this paper introduces \textbf{HiKE}: \textbf{Hi}erarchical \textbf{K}orean-\textbf{E}nglish code-switching benchmark consisting of $1,121$ high-quality CS utterances covering various topics (e.g., Software Engineering, Language Education).
To reflect the various forms of CS that occur in real-world scenarios, we labeled the utterances according to three hierarchical CS-levels (i.e., word-, phrase- and sentence-level), as illustrated in \autoref{fig:motivation}. Using HiKE, we conduct a two-part analysis. First, we evaluate ten multilingual ASR models, spanning a range of architectures and model sizes, to assess their Korean-English CS-ASR capabilities. Second, we compare the CS-ASR performance achieved by fine-tuning with two different data types: natural word- and phrase-level CS data and synthesized sentence-level CS data.

The contributions of this paper are three-fold: 
\textbf{First,} to the best of our knowledge, HiKE is the first to release a publicly available non-synthetic Korean-English CS speech recognition benchmark, which includes loanword labels and CS-level annotations.
\textbf{Second,} leveraging our loanword and hierarchical CS-level annotations, we precisely measure how the performance of $10$ multilingual ASR models varies depending on the type of CS.
\textbf{Third,} our fine-tuning experiments demonstrate that a model's CS-ASR capabilities can be effectively enabled through fine-tuning, and that this is achievable not only with natural CS data but also with synthetic data created by concatenating monolingual utterances.

\begin{table}[t!]
\centering
\resizebox{\linewidth}{!}{%
\begin{tabular}{c|ccc|cc}
\toprule
\midrule
\multirow{2}{*}{\textbf{Topic}} & \multicolumn{3}{c|}{\textbf{CS-Level}} & \multirow{2}{*}{\textbf{Total}} & \multirow{2}{*}{\textbf{Proportion}} \\
\cmidrule{2-4}
&  \textbf{word} & \textbf{phrase} & \textbf{sentence} \\
\midrule
\textbf{academic}               & 40 & 110 & 7   & 157 & 14.0\% \\
\textbf{business}               & 45 & 112 & 12  & 169 & 15.1\% \\
\textbf{entertainment}          & 53 & 29  & 8   & 90  & 8.0\% \\
\textbf{everyday conversation}  & 69 & 76  & 13  & 158 & 14.1\% \\
\textbf{language education}     & 82 & 75  & 1   & 158 & 14.1\% \\
\textbf{medical}                & 28 & 48  & 0   & 76  & 6.8\% \\
\textbf{software development}   & 43 & 113 & 6   & 162 & 14.5\% \\
\textbf{travel and culture}     & 97 & 44  & 10  & 151 & 13.5\% \\
\midrule
\textbf{Total}                  & 457 & 607 & 57 & \textbf{1,121} & \textbf{100\%} \\
\midrule
\bottomrule
\end{tabular}%
}
\caption{
    \textbf{Number of Utterances by Topic \& CS-Level}
  }
\label{tab:data_stat_topic}
\end{table}

\section{Related Work}
\label{sec:related_works}

\subsection{Automatic Speech Recognition}
Early deep learning-based ASR systems were often streaming-based, using architectures such as Connectionist Temporal Classification (CTC) to make predictions on short audio frames~\cite{wav2vec2, hubert, sensevoice}. A subsequent paradigm shift occurred towards non-streaming, Transformer-based encoder-decoder models~\cite{transformer, whisper, seamless}, which leverage full audio context to achieve superior accuracy and robustness. The most recent trend involves directly leveraging pretrained large language models (LLMs)~\cite{gemma3n, audioflamingo3, gpt4o}, offering high accuracy and task extensibility, but requiring immense computational resources.

\subsection{Code-Switching Speech Datasets}
\label{sec:cs-speech-datasets}
Publicly available CS-ASR datasets are exceptionally rare due to the inherent challenges of data collection. Even for Mandarin-English, whose large speaker population has resulted in a relative abundance of data~\cite{ASRU, CS-Dialogue, DOTA-ME-CS}, only a handful of public benchmarks exist. The scarcity of datasets is particularly acute for typologically distant language pairs like Korean-English, presenting a critical bottleneck for research into improving CS-ASR performance. While a government-funded Korean dataset~\cite{aihub_cs_data} exists, its use is restricted to Korean nationals, making it inaccessible to the global research community. 

To address the scarcity of Code-Switching data, previous research has explored methods such as concatenating monolingual utterances to generate CS samples \cite{hussein2024speech, 8462180} or utilizing Text-to-Speech (TTS) models to produce synthetic CS data \cite{yan2025cs, yu2023code, Sharma2020ImprovingLR} for the training and evaluation of Automatic Speech Recognition (ASR) systems. However, concatenated data often suffers from mid-utterance discontinuities in vocal traits (e.g., gender and recording environment) and is restricted to sentence-level CS. Similarly, TTS-generated data may exhibit poor audio quality and, due to its synthetic nature, fails to capture the diverse CS patterns and acoustic variability found in real-world environments, making it unsuitable for evaluating performance in authentic scenarios. To the best of our knowledge, HiKE is the first globally accessible non-synthetic Korean-English CS benchmark.

\section{HiKE}
\label{sec:dataset}

\begin{table*}[t!]
\centering
\resizebox{\textwidth}{!}{%
\begin{tabular}{l|c|cccc|cccc|cc}
\toprule
\midrule
\multirow{2}{*}{\textbf{}} & \multirow{2}{*}{\textbf{\# Params}} & \multicolumn{4}{c|}{\textbf{Mixed Error Rate (MER)}} & \multicolumn{4}{c|}{\textbf{Point of Interest Error Rate(PIER)}} & \multicolumn{2}{c}{\textbf{Monolingual}} \\
\cmidrule(lr){3-6} \cmidrule(lr){7-10} \cmidrule(lr){11-12}
& & \textbf{Word} & \textbf{Phrase} & \textbf{Sentence} & \textbf{Overall} & \textbf{Word} & \textbf{Phrase} & \textbf{Sentence} & \textbf{Overall} & \textbf{KOR} & \textbf{ENG} \\
\midrule
\textsc{SenseVoice-Small} \cite{sensevoice}     & 234M & 27.4 & \underline{36.4} & \textbf{23.6} & 32.5 & 52.2 & \underline{56.5} & \textbf{37.1} & 54.7 & 6.4 & 7.6 \\
\midrule
\textsc{Whisper-Tiny} \cite{whisper}            & 38M & 73.7 & \underline{111.4} & \textbf{36.2} & 93.6 & \underline{81.7} & 77.9 & \textbf{66.9} & 78.9 & 11.9 & 14.6 \\
\textsc{Whisper-Base} \cite{whisper}            & 74M & \underline{116.5} & 81.8 & \textbf{25.4} & 91.2 & \underline{90.2} & 73.2 & \textbf{38.7} & 78.1 & 7.8 & 9.8 \\
\textsc{Whisper-Small} \cite{whisper}           & 244M & \underline{50.6} & 41.3 & \textbf{19.9} & 43.5 & \underline{58.0} & 46.7 & \textbf{31.5} & 50.1 & 4.5 & 8.3 \\
\textsc{Whisper-Medium} \cite{whisper}          & 769M & \underline{39.1} & 27.9 & \textbf{16.8} & 31.3 & 41.9 & \underline{41.3} & \textbf{30.6} & 41.3 & 3.4 & 4.6 \\
\textsc{Whisper-Large} \cite{whisper}           & 1.5B & \underline{28.5} & 25.3 & \textbf{20.0} & 26.1 & \underline{45.6} & 31.3 & \textbf{25.8} & 36.0 & 3.2 & 4.4 \\
\textsc{Seamless-M4T-v2-large} \cite{seamless}  & 2.3B & \underline{108.8} & 70.1 & \textbf{61.0} & 83.6 & \underline{72.4} & 60.9 & \textbf{57.3} & 64.7 & 6.4 & 6.3 \\
\midrule
\textsc{Gemma-3n} \cite{gemma3n}                & 8B & \underline{99.6} & 64.6 & \textbf{54.3} & 76.6 & \underline{69.2} & 47.8 & \textbf{40.3} & 54.8 & 10.7 & 13.0 \\
\textsc{Audio-Flamingo-3} \cite{audioflamingo3}     & 8.3B & \textbf{78.5} & \underline{82.2} & 79.4 & 80.7 & 104.1 & \textbf{97.8} & \underline{112.9} & 100.2 & 25.1 & 8.8 \\
\textsc{GPT-4o-transcribe} \cite{gpt4o}     & N/A & \textbf{15.6} & 25.0 & \underline{28.3} & 21.8 & \textbf{25.8} & 30.2 & \underline{32.3} & 28.8 & 2.5 & 3.3 \\
\midrule
\bottomrule
\end{tabular}%
}   
\caption{
\textbf{Benchmark Results.} For each model, the \textbf{best} and \underline{worst} scores are bolded and underlined, respectively. Monolingual performance is measured on the FLEURS dataset, using CER for Korean and WER for English.
}
\label{tab:bench_result}
\end{table*}

\subsection{Script Writing \& Cloning}
We built our dataset through a human-LLM collaborative process to ensure both high quality and minimal human effort. We began by manually authoring $575$ seed scripts across $8$ topics (\autoref{tab:data_stat_topic}). Each script was then used as a one-shot example to prompt \textsc{Claude-3.5-Sonnet}~\cite{claude} to generate a new script mimicking the original's topic and CS structure, utilizing the prompt detailed in \autoref{sec:clone_prompt}. To prevent excessive similarity among the generated scripts and maintain a balance between data volume and diversity, we opted to generate only one clone per original script. As a final quality control step, all generated scripts were manually reviewed and corrected by the authors.

\subsection{Recording}
\label{sec:recording}
We recruited $13$ bilingual Korean-English speakers, each of whom recorded $50$ to $100$ scripts in a quiet environment using a web-based interface on their personal devices, such as laptops and mobile devices. This process yielded an initial batch of $1,150$ recordings. Following a manual review by the authors, $29$ samples that deviated from the script were discarded, resulting in a final curated dataset of $1,121$ high-quality utterances totaling approximately $2.2$ hours.

\subsection{Metrics}
\label{sec:metrics}
Given that CS-ASR involves mixing languages with different linguistic properties, many prior works~\cite{ASRU, CS-Dialogue, DOTA-ME-CS} have adopted the Mixed Error Rate (MER), which evaluates character-based languages such as Mandarin and Korean at the character level and word-based languages such as English at the word level. In contrast to MER, which assesses the entire utterance, the Point of Interest Error Rate (PIER)~\cite{PIER} was later proposed to specifically evaluate performance at the points where language transitions occur. In this paper, we evaluate the CS-ASR capabilities of models using both MER and PIER. For our PIER evaluation, we tagged the words at the locations where code-switching occurs. This allowed us to assess whether the ASR model can accurately switch languages precisely at these transition points.

\subsection{Hierarchical CS-Level}
In contrast to previous work~\cite{ASRU, ascend, talcs} that only classified CS by its location within the utterance (inter- vs. intra-sentential), we propose a more granular, three-level classification to analyze how models handle intra-sentential CS between typologically distant languages.
HiKE categorizes code-switching into three distinct levels: sentence, word, and phrase. Sentence-level CS is the most predictable, as switches occur only at utterance boundaries. Word-level CS involves the substitution of single lexical units, primarily testing a model's bilingual lexicon. Phrase-level CS, in contrast, poses a more complex grammatical challenge, as it can introduce irregular structures like altered word order, a difficulty that is significantly amplified for typologically distant pairs such as Korean and English.

\subsection{Loanwords Post-processing}
Loanwords are words adopted from a foreign language and adapted to the phonology and orthography of the new language. For example, the Korean loanword `버스' [b\textschwa s] and the English word `bus' [b\textturnv s] are pronounced almost identically and can be used interchangeably in a CS context. This creates an evaluation challenge: If a ground-truth label is strictly constrained to either Hangul or the Roman alphabet, a model producing the alternate—yet perfectly valid—transcription would be unfairly penalized. This ambiguity introduces significant noise into the assessment of CS performance. To avoid this problem, we meticulously labeled all loanwords contained in our dataset. Subsequently, during evaluation, both the Korean and English versions of these loanwords were treated as valid answers. Through our loanword labeling process, we achieved a more precise evaluation by decreasing measurement noise by an average of $4.9\%$ in MER and $7.9\%$ in PIER.

\begin{table*}[t!]
\centering
\resizebox{\textwidth}{!}{%
\begin{tabular}{l|cccc|cccc|cc}
\toprule
\midrule
\multirow{2}{*}{\textbf{}} & \multicolumn{4}{c|}{\textbf{Mixed Error Rate (MER)}} & \multicolumn{4}{c|}{\textbf{Point of Interest Error Rate (PIER)}} & \multicolumn{2}{c}{\textbf{Monolingual}} \\
\cmidrule(lr){2-5} \cmidrule(lr){6-9} \cmidrule(lr){10-11}
& \textbf{Word} & \textbf{Phrase} & \textbf{Sentence} & \textbf{Overall} & \textbf{Word} & \textbf{Phrase} & \textbf{Sentence} & \textbf{Overall} & \textbf{KOR} & \textbf{ENG} \\
\midrule
\textsc{Whisper-Medium}                 & 39.1 & 27.9 & 16.8 & 31.3 & 41.9 & 41.3 & 30.6 & 41.3 & 3.4 & 4.6 \\
\midrule
\textbf{(a)} FT with Natural Intra-Sentential CS Data       & \textbf{8.3\diff{-30.8}}   & \textbf{9.6\diff{-18.4}} & 7.4\diff{-9.4} & \textbf{9.0\diff{-22.3}} & \textbf{18.6\diff{-23.3}} & \textbf{19.9\diff{-21.4}} & 21.0\diff{-9.7} & \textbf{19.5\diff{-21.8}} & 6.0\diff{+2.6} & 5.2\diff{+0.6} \\
\textbf{(b)} FT with Synthetic Inter-sentential CS Data       & 19.1\diff{-20.0}    & 25.6\diff{-2.3} & 5.8\diff{-11.0}   & 22.1\diff{-9.2} & 33.7\diff{-8.2} & 35.6\diff{-5.7}  & 14.5\diff{-16.1} & 34.5\diff{-6.8} & 3.7\diff{+0.3} & 5.1\diff{+0.5} \\
\textbf{(a) + (b)} FT with Both Data    & 20.7\diff{-18.4}    & 21.7\diff{-6.3} & \textbf{4.8\diff{-11.9}}  & 20.4\diff{-11.0}  & 27.3\diff{-14.6} & 37.6\diff{-3.7}  & \textbf{11.3\diff{-19.4}} & 33.6\diff{-7.7} & 3.9\diff{+0.5} & 4.9\diff{+0.3} \\
\midrule
\bottomrule
\end{tabular}%
}
\caption{
    \textbf{Fine-Tuning (FT) Results.} For each metric, the \textbf{best} scores are highlighted in bold. Monolingual performance is measured on the FLEURS dataset, using CER for Korean and WER for English.
  }
\label{tab:fine-tuning_result}
\end{table*}

\section{Experiments}
\subsection{CS-ASR of Multilingual ASR Models}
We evaluated $10$ multilingual ASR models with diverse architectures (CTC, Transformer, and LLM-based), assessing their CS-ASR performance using MER and PIER metrics. As shown in \autoref{tab:bench_result}, a severe performance drop occurred on CS data across all models, with the MER increasing by a factor of $3$ to $13$ compared to monolingual performance, revealing significant limitations for practical use. A closer look at specific architectures reveals further nuances. For instance, the CTC-based \textsc{SenseVoice-Small}, despite showing stable performance across CS-levels and a better overall MER than the similarly-sized \textsc{Whisper-Small}, still exhibited a high error rate at the actual code-switching points. This suggests that while CTC architectures may be robust in overall transcription, they struggle specifically with the precise moment of language transition.

In contrast, models leveraging Large Language Models (LLMs) trained on extensive text corpora displayed distinct behavioral patterns compared to speech-centric models. Specifically, despite having over five times the parameter count of \textsc{Whisper-Large}, \textsc{Gemma-3n} and \textsc{Audio-Flamingo-3} failed to achieve competitive results in CS-ASR, with both MER and PIER exceeding $54$. \textsc{GPT-4o-Transcribe} was the only LLM-based model to outperform \textsc{Whisper-Large}. Notably, while most non-LLM models performed best on sentence-level CS and worst on word-level CS, \textsc{GPT-4o-Transcribe} exhibited the opposite trend, achieving its highest performance on word-level CS and its weakest on sentence-level CS. We hypothesize that this reflects the distribution of its training data, as large text corpora are rich in word-level CS but contain comparatively few instances of sentence-level CS.


To analyze the effect of model scale on CS-ASR performance, we evaluated the Whisper family of models. The results show a clear correlation between size and capability: CS-ASR performance, nearly absent in the smallest models (e.g., Tiny and Base), gradually emerges with increasing scale. Despite this trend, even the largest model's error rate on CS data was over five times higher than on monolingual data, indicating that model scaling alone is an insufficient solution for achieving practical CS-ASR performance.

\subsection{Fine-Tuning with a Synthetic CS Dataset}
\subsubsection{Experimental Details}
To investigate the effect of fine-tuning with CS-ASR performance, we prepared two distinct types of training data. The first was a natural, intra-sentential (word- and phrase-level) CS dataset from AIHub~\cite{aihub_cs_data}. The second was a synthetic sentence-level CS dataset, which we generated by concatenating monolingual Korean and English utterances from the FLEURS~\cite{FLEURS} and Common Voice~\cite{commonvoice} datasets. Using these, we fine-tuned \textsc{Whisper-Medium}~\cite{whisper} under three conditions: using the intra-sentential CS set only, the synthetic inter-sentential set only, and a combination of both.

\subsubsection{Results}
The results in \autoref{tab:fine-tuning_result} reveal that fine-tuning is an effective method for enabling CS-ASR, and that this can be achieved not only with natural intra-sentential CS data \textbf{(a)} but also with synthetic inter-sentential CS data \textbf{(b)} created by concatenating monolingual utterances.
As shown in row \textbf{(b)} of \autoref{tab:fine-tuning_result}, fine-tuning of synthetic CS data alone improved both the overall MER and PIER by more than $6.8\%$. Given the wide availability and relative ease of collecting monolingual data compared to authentic CS data, this finding suggests that data synthesis via utterance concatenation represents a promising and cost-effective direction for training CS-ASR models, especially in resource-constrained scenarios.
However, fine-tuning on natural intra-sentential CS data \textbf{(a)} yielded greater performance improvements than fine-tuning on synthetic inter-sentential data \textbf{(b, c)} across almost all metrics; the only exceptions were the MER and PIER for sentence-level CS. Although this result indicates that it is currently preferable to use natural intra-sentential CS data for fine-tuning when available, we anticipate that these limitations can be overcome. With the development of more sophisticated data synthesis or fine-tuning techniques, it may become possible to train robust CS-ASR models using only synthetic data.

\subsection{Qualitative Results}
\begin{figure}
    \centering
    \includegraphics[width=\linewidth]{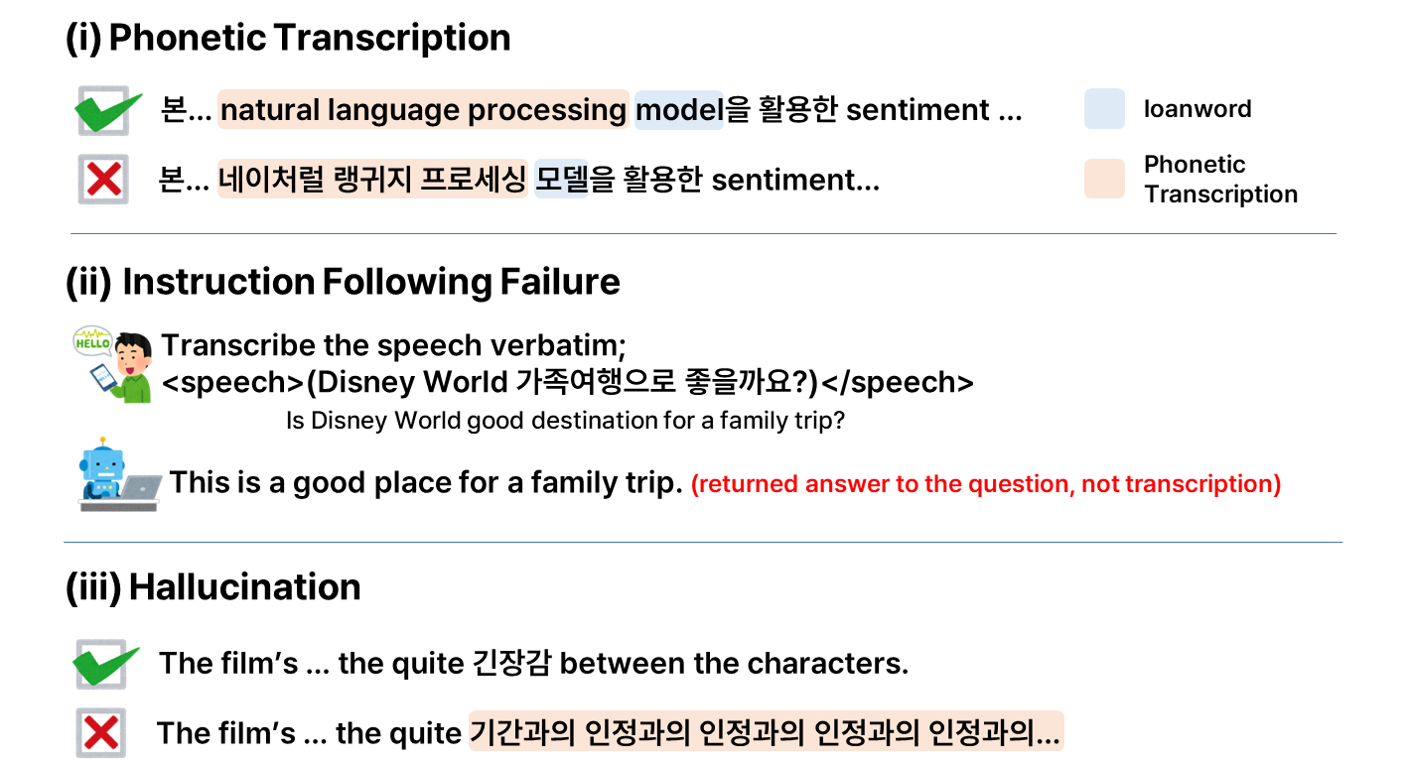}
    \caption{\textbf{CS-ASR Fail Cases}}  
    \label{fig:fail_cases}
\end{figure}

In our analysis of the CS transcription results, we primarily observed three types of errors, as illustrated in \autoref{fig:fail_cases}.
\textbf{(i)} Phonetic Transcription refers to cases where, for words that are not loanwords, the model does not transcribe them in the correct language but instead writes them out phonetically using the script of the other language. This error was commonly observed across all models. \textbf{(ii)} Instruction Following Failure is an error that occurs in multi-task models (e.g., \textsc{Whisper}) capable of handling tasks beyond transcription, such as translation and question answering. This was especially pronounced with \textsc{Audio-Flamingo-3}; while this error was rare in monolingual settings, its high frequency in CS environments made the model unreliable for transcription without a separate verification step. Finally, \textbf{(iii)} Hallucination is an error common in seq2seq models, including those based on Transformers and LLMs. It refers to cases where the model incorrectly generates repetitive or excessive content that is not present in the audio.

Qualitatively, while errors like phonetic transcription often occurred even with monolingual data, instruction following failures and hallucinations increased markedly in CS data. We attribute this trend to the fact that during training, ASR models are exposed to abundant monolingual data but extremely rare CS data, which hinders the generalization of their performance to CS scenarios.

\section{Conclusion}
In this paper, we propose HiKE, the first public high-quality hierarchical benchmark for Korean-English CS-ASR with hierarchical CS-level labels and loanword labels. Our evaluation of $10$ multilingual models in HiKE shows that strong monolingual performance does not generalize to CS scenarios. Specifically, we found that models are less accurate on word- and phrase-level CS, which feature dense, irregular switch points and complex grammatical structures, in contrast to the more predictable sentence-level CS. Furthermore, our fine-tuning experiments demonstrate that a model's CS-ASR capabilities can be improved, even with synthetic inter-sentential CS data created by concatenating monolingual utterances.
We believe that this work will serve as a foundation for future research in CS-ASR, including developing models for diverse language pairs beyond Korean-English, generating high-quality synthetic CS data, and analyzing the generalization of CS-ASR capabilities.

\section*{Limitations}
Our study has two primary limitations. First, the scarcity of ASR models that support both Korean and English precluded a broad comparative study of diverse architectures. This was especially true for LLM-based models, as only \textsc{GPT-4o-Transcribe} yielded a meaningful CS-ASR performance, which prevented a reliable analysis of their common characteristics. Second, the scarcity of high-quality CS data restricted our fine-tuning experiments to a small scale, precluding a thorough investigation into methods for eliciting robust CS capabilities, such as by scaling up synthetic data. We believe these limitations point to several avenues for future research.

\section*{Acknowledgments}
We appreciate Hyunwoo Kim for proofreading and the anonymous reviewers for their insightful feedback and suggestions. Special thanks to DongChan Shin (@DongChanS) for pointing out issues in the preprocessing pipeline on GitHub. We also thank the Theta One Team for supporting our work.

This work was supported by the Tech Incubator Program for Startup Korea (RS-2024-00507331) funded by the Ministry of SMEs and Startups (MSS, S. Korea).

\newpage
\bibliography{custom}

\newpage
\appendix

\begin{figure}
    \centering
    \includegraphics[width=\linewidth]{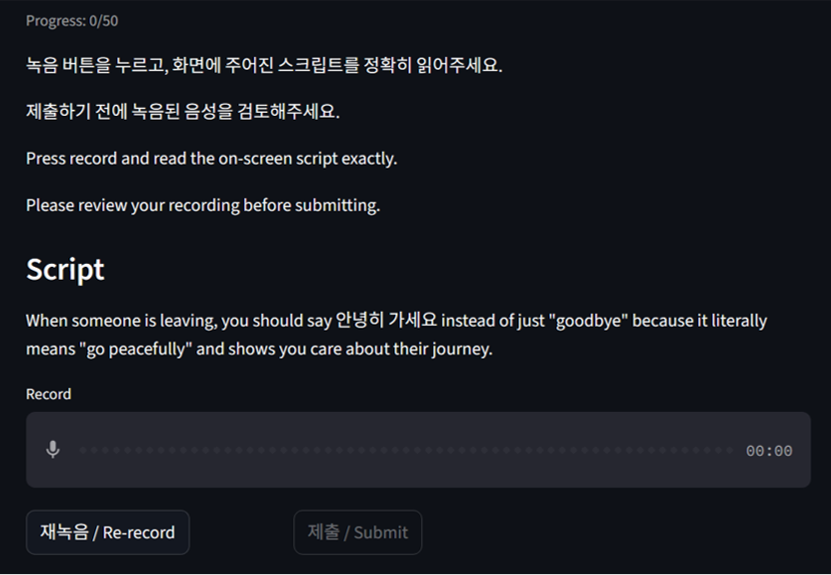}
    \caption{\textbf{Screenshot of Recording Tool}} 
    \label{fig:recording_tool}
\end{figure}

\begin{table}[htbp]
\centering
\begin{tabular}{lc}
\toprule
\textbf{Characteristic} & \textbf{Count} \\ \toprule
\textit{Gender}         &                \\ \bottomrule
Male                    & 11             \\
Female                  & 2              \\ \toprule
\textit{Educational Level} &             \\ \bottomrule
Undergraduate           & 6              \\
Bachelor's              & 7              \\ \bottomrule
\textbf{Total}                   & 13             \\ \bottomrule
\end{tabular}
\caption{\textbf{Demographics of Participants}}
\label{tab:statistics}
\end{table}

\begin{figure*} 
    \centering 
    \fbox{ 
        \begin{minipage}{0.95\textwidth} 
            Referring to the example below, create a new Korean-English code-switching sample.
First, analyze the example to identify its characteristics and themes, then return a new code-switching sample that reflects those characteristics, wrapped in a <sample> tag.

<example>

\textcolor{blue}{\{example\}}

</example>
        \end{minipage}
    }
    \caption{\textbf{Script Cloning Prompt}}
    \label{fig:clone_prompt}
\end{figure*}

\section{Experimental Setup}
Unless otherwise specified, all experiment results were obtained from a single run on a NVIDIA RTX 6000 Ada GPU. We used pyTorch 2.8.0 and transformers 4.56.2 for our experiments.

In finetuning experiments, all models were fine-tuned for approximately 1 epoch on a single A100 GPU with a batch size of $16$, using a cosine annealing scheduler with a $10\%$ warmup and a peak learning rate of $1e-5$.

\section{Dataset Recording}
\label{sec:recording_detail}
As mentioned in \autoref{sec:recording}, participants performed the data recordings using a web-based interface like the one shown in \autoref{fig:recording_tool}. The webpage provided recording instructions and scripts in both Korean and English, and allowed participants to review their own recordings and perform re-takes if necessary. The demographic distribution of the participants, specifically regarding gender and educational background, is summarized in \autoref{tab:statistics}.

\section{License}
Our experiments utilize the official code implementation of PIER~\cite{PIER} (Apache 2.0 License), along with other standard machine learning libraries. Our own evaluation code, developed for the HiKE benchmark, will also be publicly released under the Apache 2.0 License.

\section{Script Cloning Prompt}
\label{sec:clone_prompt}
\autoref{fig:clone_prompt} shows the prompt used to clone a new script based on a given original script.

                                                                                                                                                    \section{CS-Level Determination}
\label{sec:cs-level-determination}
To ensure consistent and automated determination of CS-levels, we adopted the following hierarchical procedure. First, a sample is classified as sentence-level code-switching if it contains at least one sentence composed entirely of English and another entirely of Korean. Among the remaining samples, those containing a sequence of two or more consecutive words in a language different from the rest are categorized as phrase-level code-switching. Finally, any remaining samples where the code-switching consists of only a single, non-consecutive word are classified as word-level code-switching. \autoref{fig:determination_algorithm} presents the pseudocode for this process. $L_{main}$ represents the main language of the utterance annotated by the authors, whereas $L_{other}$ indicates the other language.

\begin{figure*}[ht]
\centering
\begin{minipage}{0.95\textwidth}
\begin{algorithm}[H]
\caption{CS-Level Determination Procedure}
\begin{algorithmic}[1] 
\Procedure{Level-Detecting}{$RawText, L_{main}, L_{other}$}
    \State 
    \Comment{Step 1 \& 2: Pre-processing \& Mapping}
    \State $CleanText \gets \text{RemoveSpecialChars}(RawText)$
    \State $Seq \gets \text{Map characters to } \{L_{main}, L_{other}\}$
    \State $Segments \gets \text{Split}(Seq, \text{delimiters}=\{., !, ?\})$
    
    \State \Comment{Step 3: Level Checking (Independent Detection)}
    \For{each $segment$ in $Segments$}        
        \If{$segment$ consists \textbf{only} of $L_{other}$}
            \State $has\_sentence \gets \text{true}$
        \EndIf
        
        \If{$segment$ contains a sequence of $(L_{other}, L_{other})$}
            \State $has\_phrase \gets \text{true}$
        \EndIf
        
        \If{$segment$ contains an isolated unit of $L_{other}$}
            \State $has\_word \gets \text{true}$
        \EndIf
    \EndFor
    
    \State \Comment{Step 4: Priority-based Classification}
    \If{$has\_sentence$} \State \Return "sentence"
    \ElsIf{$has\_phrase$} \State \Return "phrase"
    \ElsIf{$has\_word$} \State \Return "word"
    \Else \State \Return "None"
    \EndIf
\EndProcedure
\end{algorithmic}
\end{algorithm}
\end{minipage}
\caption{CS-Level Determination Procedure}
\label{fig:determination_algorithm}
\end{figure*}

\end{document}